\documentclass[conference]{IEEEtran}
\IEEEoverridecommandlockouts
\usepackage[hyphens]{url}
\usepackage[hidelinks]{hyperref}
\hypersetup{breaklinks=true}
\usepackage{cite}
\usepackage{amsmath,amssymb,amsfonts}
\usepackage{graphicx}
\usepackage{subcaption}
\usepackage{textcomp}
\usepackage{xcolor}
\usepackage{url}
\usepackage{float}
\def\BibTeX{\rm B\kern-.05em{\sc i\kern-.025em b}\kern-.08em
    T\kern-.1667em\lower.7ex\hbox{E}\kern-.125emX}

\usepackage{algorithm}
\usepackage{algorithmicx}
\usepackage[noend]{algpseudocode}

\begin{document}

\title{Autoencoder-based Anomaly Detection in Streaming Data with Incremental Learning and Concept Drift Adaptation\\
\thanks{This paper was supported in part by funding from the European Research Council (ERC) under grant agreement No. 951424 (Water Futures) and the European Union's Horizon 2020 research and Innovation programme under grant agreement No. 739551 (KIOS Co).}
}

\author{
	\IEEEauthorblockN{
		Jin Li\textsuperscript{1, 2},
		Kleanthis Malialis\textsuperscript{1},
		Marios M. Polycarpou\textsuperscript{1, 2}
	}
	\IEEEauthorblockA{
		\textsuperscript{1} \textit{KIOS Research and Innovation Center of Excellence}\\
		\textsuperscript{2} \textit{Department of Electrical and Computer Engineering}\\
        University of Cyprus, Nicosia, Cyprus\\
		\{li.jin, malialis.kleanthis, mpolycar\}@ucy.ac.cy
		\\ORCID: \{0000-0002-3534-524X, 0000-0003-3432-7434, 0000-0001-6495-9171\}
	}
}

\maketitle
\begin{abstract}
In our digital universe nowadays, enormous amount of data are produced in a streaming manner in a variety of application areas. These data are often unlabelled. In this case, identifying infrequent events, such as anomalies, poses a great challenge. This problem becomes even more difficult in non-stationary environments, which can cause deterioration of the predictive performance of a model. To address the above challenges, the paper proposes an autoencoder-based incremental learning method with drift detection (strAEm++DD). Our proposed method strAEm++DD leverages on the advantages of both incremental learning and drift detection. We conduct an experimental study using real-world and synthetic datasets with severe or extreme class imbalance, and provide an empirical analysis of strAEm++DD. We further conduct a comparative study, showing that the proposed method significantly outperforms existing baseline and advanced methods.
\end{abstract}


\section{Introduction}

In recent years, we are witnessing an explosive growth in the availability of streaming data in various application areas. The deployment of online learning algorithms for streaming data in the real world is restricted by some key challenges, including class imbalance, label unavailability and nonstationary environments.

\textbf{Class imbalance}: The anomaly detection problem can be considered as an extreme case of class imbalance problem, where severe or extreme imbalance exists. The existence of class imbalance can dramatically deteriorate the performance of classifiers by introducing a biased prediction for the majority class \cite{krawczyk2016learning}. When dealing with streaming data, this issue becomes even more complex.

\textbf{Label unavailability}:
Most anomaly detection systems adopt signature-based methods or data mining-based methods which rely on labelled training data. \cite{eskin2002geometric}. Labeled data are typically expensive to acquire in real-time applications, or might even be unavailable.

\textbf{Nonstationary environments}: Often, we assume that the process generating the stream data is stationary. However, in many real cases, the generating process is characterized by an intrinsic nonstationary phenomenon, which is also referred to as concept drift. Various factors can trigger concept drift, such as, seasonality or periodicity effects (e.g., consumption in a water distribution network), incipient faults (e.g., in monitoring of critical infrastructure systems), changes in users' interests or preferences (e.g., in recommendation systems), and changes in users' behaviour (e.g., in fraud prevention or network intrusion detection systems).

To deal with the above challenges, a predictive model should ideally have the following properties: (i) be able to effectively identify infrequent events, i.e., anomalies; (ii) be able to adapt to concept drift to maintain high performance in nonstationary environments; (iii) be able to learn with unlabelled data since labelled data may not accessible in some real use cases. The two key contributions of this work are as follows:
\begin{enumerate}
	\item We develop strAEm++DD, an autoencoder (AE)-based incremental learning algorithm with a concept drift detection mechanism. strAEm++DD satisfies the aforementioned properties as it can effectively identify infrequent events, detect concept drift, and does not rely on supervision. One of its novel characteristics is its explicit concept drift detection module, which works in a synergistic manner with incremental learning for effective adaption to nonstationary environments.

	\item We conduct an experimental study using real-world and synthetic datasets with severe or extreme class imbalance and provide an empirical analysis of strAEm++DD. We also conduct a comparative study, showing that the proposed method significantly outperforms existing baseline and advanced methods, which demonstrates its effectiveness in addressing the above challenges. 
\end{enumerate}

The paper is organised as follows. Section~\ref{sec:background} provides the background material, necessary to understand the contributions of this paper. Related work is described in Section~\ref{sec:related}. The proposed method is discussed in Section~\ref{sec:method}. The experimental setup is described in Section~\ref{sec:exp_setup}. The empirical analysis of the proposed approach and comparative studies of learning methods are provided in Section~\ref{sec:exp_results}. Some concluding remarks are discussed in Section~\ref{sec:conclusion}.

\section{Online Learning}\label{sec:background}

\textbf{Online} learning considers a data generating process that provides at each time step $t$ a batch of examples $S = \{B^t\}_{t=1}^T$, where each batch is defined as $B^t = \{(x^t_i,y^t_i)\}^M_{i=1}$. The total number of steps is denoted by $T \in [1, \infty)$ where the data are typically sampled from a long, potentially infinite, sequence. The number of examples at each step is denoted by $M$. If $M=1$, it is termed \textbf{one-by-one online} learning, otherwise it is termed \textbf{batch-by-batch online} learning \cite{ditzler2015learning}. This work focuses on one-by-one learning; i.e., $B^t = (x^t, y^t)$, which is important for real-time monitoring. The examples are drawn from an unknown time-varying probability distribution $p^{t}(x,y)$, where $x^t \in \mathbb{R}^d$ is a $d$-dimensional vector in the input space $X \subset \mathbb{R}^d$, $y^t \in \{1, ..., K\}$ is the class label and $K \geq 2$ is the number of classes. This work's focus is on anomaly detection, therefore, the number of classes is $K=2$ (``normal'', ``anomalous''). 

A one-by-one online classifier receives a new instance $x^t$ at time $t$ and makes a prediction $\hat{y}^t$ based on a concept $h: X \to Y$. In \textbf{online supervised} learning, the classifier receives the true label $y^t$, its performance is evaluated using a loss function and is then trained based on the loss incurred. The process is repeated at each time step. The continual adaptation of the model without complete re-training, i.e., $h^t = h^{t-1}.train(\cdot)$ is termed \textbf{incremental} learning \cite{losing2018incremental}.

In data streaming applications, however, the class label cannot be typically provided in real-time. To alleviate this problem, the community has turned to alternative learning paradigms, such as, \textbf{online semi-supervised} learning \cite{dyer2013compose} which initially relies on a small fraction of labelled data, and \textbf{online active} learning \cite{zliobaite2013active, malialis2022nonstationary, malialis2022augmented, malialis2020data} which deals with strategies that learn to query a human expert for ground truth information (e.g., class labels) of selected examples. Despite their effectiveness, the above paradigms still rely on labelled data. This work focuses on one-by-one \textbf{online unsupervised} learning, where no class labels are required, i.e., $B^t = (x^t)$.

A major challenge encountered in some streaming applications is the presence of \textbf{class imbalance} \cite{he2008learning}. It occurs when at least one class is under-represented, thus constituting a minority class. In binary classification, imbalance is defined as follows:
\begin{equation}
\exists y_0, y_1 \in Y \quad p^t(y=y_0) >> p^t(y=y_1),
\end{equation}
\noindent where $y_1$ represents the minority class.

Another significant challenge encountered in some streaming applications is that of data \textbf{nonstationarity} \cite{ditzler2015learning, gama2014survey}, typically caused by \textbf{concept drift}, which represents a change in the joint probability. The drift between steps $t_i$ and $t_j$, where $i \ne j$, is defined as follows:
\begin{equation}
\quad p^{t_i}(x,y) \neq p^{t_j}(x,y)
\end{equation}

There are three types of concept drift: (i) a change in prior
probability $p(y)$ (ii) a change in class-conditional probability
or likelihood $p(x|y)$ and (iii) a change in posterior probability
$p(y|x)$.

Learning in the presence of both drift and imbalance, even in supervised settings, remains an open research challenge \cite{wang2018systematic}.

\section{Related Work}\label{sec:related}

\subsection{Concept drift adaptation}
Approaches dealing with concept drift are, typically, categorised as passive or active \cite{ditzler2015learning}.

\subsubsection{Passive methods}
In this category, methods can be divided into memory-based and ensembling methods. A memory-based algorithm uses a memory component to maintain a set of recent examples on which the classifier is trained; for example, \cite{widmer1996learning} uses an adaptive sliding window. 
Ensembling \cite{krawczyk2017ensemble} is made up with a set of classifiers which can be either added or removed depending on their performance. A representative method is DDD\cite{minku2011ddd}. A series of research studies have been proposed to deal with imbalanced data in nonstationary environments, such as QBR\cite{malialis2018queue}, AREBA\cite{malialis2020online}, and ROSE\cite{cano2022rose}. The vast majority of incremental learning methods rely on supervision.

\subsubsection{Active methods}
They rely on an explicit detection of the change in the data distribution to activate an adaptation mechanism\cite{ditzler2015learning}. Two categories of detection mechanisms are explored, statistical tests (e.g., \cite{alippi2008justI},\cite{jaworski2020concept}), and threshold-based mechanisms (e.g., \cite{gama2004learning},\cite{ menon2020concept}). Statistical tests monitor the statistical behaviour of the data produced, while threshold-based mechanisms monitor the errors produced for each prediction and compare them with a threshold. 

Autoencoders have been used as drift detectors. For instance, “Autoencoder Drift Detection” (ADD)\cite{menon2020concept} is a threshold-based method handling batch-by-batch mode data inputs, where the reconstruction error is employed as a proxy to detect concept drift of semi-stationary data.
Another autoencoder-based approach is \cite{jaworski2020concept}, aiming to detect concept drift by monitoring two cost functions respectively, the cross-entropy and the reconstruction error. The variation of these two cost functions serves as the concept drift detector. Hybrid approaches (e.g., HAREBA \cite{malialis2022hybrid}) have also been proposed to merge the advantages of both active and passive methods.

\subsection{Anomaly detection}
In anomaly detection \cite{chandola2009anomaly}, a classifier is trained on normal data to build a profile of ``normality'', and any behaviour that deviates from this, is flagged as anomalous. A lot of work exists for offline learning, such as, Local Outlier Factor (LOF) \cite{breunig2000lof} and One-Class Support Vector Machine (OC-SVM) \cite{scholkopf2001estimating}.

A popular algorithm is the Isolation Forest (iForest) \cite{liu2008isolation} which proposes a fundamentally different method that explicitly isolates anomalies instead of normal profiles. Specifically, it builds an ensemble of trees, and identifies anomalies as those instances which have short average path lengths on the trees. 

More recent methods use deep learning \cite{chalapathy2019deep}, e.g., an AutoEncoder (AE) \cite{sakurada2014anomaly} for anomaly detection. The major advantage of these methods is their ability to learn hierarchical discriminative features from data which, typically, make them more effective than traditional methods, particularly in complex problems \cite{chalapathy2019deep}.

In \cite{dong2018threaded}, the Streaming Autoencoder (SA) is proposed which uses an AE with incremental learning for online anomaly detection. The fundamental difference to the proposed method is that ours involves explicit drift detection (the ``active" part), while the other is solely incremental.

\section{The strAEm++DD Method}\label{sec:method}

The overview of the StrAEm++DD method is shown in Fig. 1. The prediction part is displayed in green colour. The system first observes the instance $x^t$ at time $t$, and the AE-based method outputs a prediction $\hat{y}^t$. The instance is then appended to a sliding window that keeps the most recent instances. Depending on whether or not a condition is satisfied (discussed below), the AE is incrementally updated, using the data in memory, which is displayed in orange colour. The proposed method is completely unsupervised as the ground truth $y^t$ is not available. We will be referring to this method as \textbf{strAEm++}. Then we incorporate explicit concept drift detection which we will be referring to it as \textbf{strAEm++DD}; this is displayed in grey color. The instance  $x^t$ is continuously appended into the drift moving window, which is used to compare the AE's reconstruction loss with the reference window for drift detection. Once a alarm flag is raised, the training window will be emptied and a new autoencoder will be created. strAEm++DD's pseudocode is shown in Algorithm 1.

\begin{figure}[t!]
	\centering
	\includegraphics[scale=0.68]{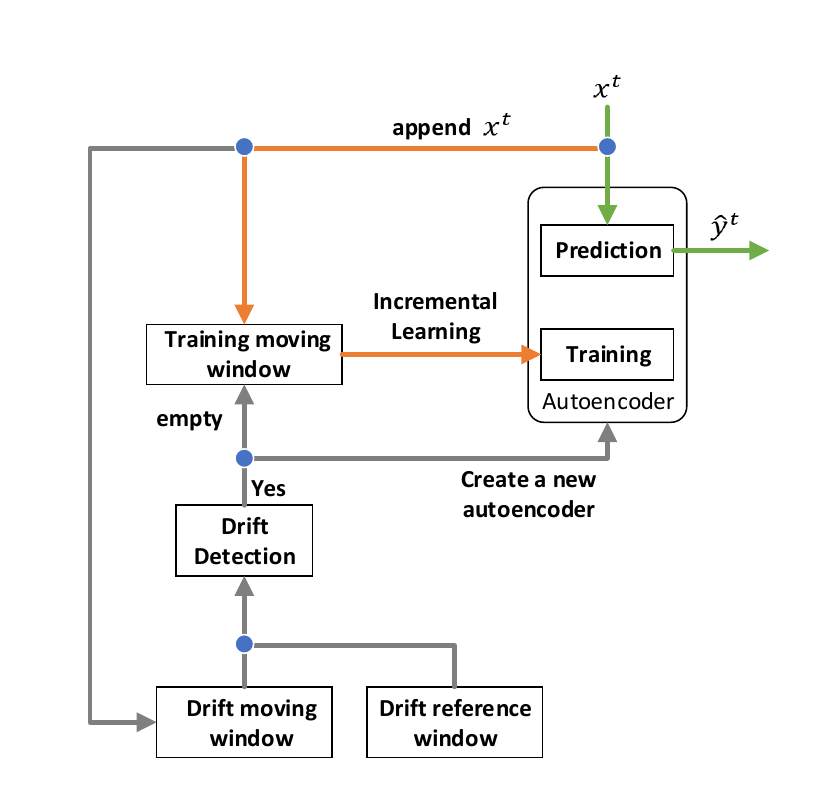}
	\caption{An overview of strAEm++DD}
	\label{fig:method}
\end{figure}

\subsection{StrAEm++}

\textbf{Model}. We consider an AE, which is a special type of a neural network that attempts to copy its input $h: X \rightarrow X$ \cite{goodfellow2016deep}. it consists of an encoder followed by a decoder and is trained to minimise the reconstruction loss between an input $x \in \mathbb{R}^d$ and its decoded version $\hat{x} = h(x) \in \mathbb{R}^d$. The reconstruction loss can be, for instance, the binary cross-entropy, as shown in Eq. (\ref{eq:ce}).

\begin{equation}\label{eq:ce}
 l(x,\hat{x}) = - \sum_d x_d \ log \ \hat{x}_d + (1 - x_d) \ log(1 - \hat{x}_d)
\end{equation}

\textbf{Memory}. The proposed method uses a sliding window $mov_{train}$ of size $W_{train}$, to store the most recent instances. At any time $t$ we maintain a queue $mov_{train}^t = \{x_t\}^t_{t-W_{train}+1}$, where for any two $x_{i}, x_{j} \in q^t$ such that $j > i$, $x_{j}$ has been observed more recently in time. The role of the sliding window is two-fold. First, it is unreasonable to assume that all arriving instances could be stored and be available at all times. Second, the sliding window implicitly addresses the problem of concept drift, as obsolete examples will eventually drop out of the queue. It is assumed that the window will be mostly populated by normal instances, due to the extreme class imbalance of an anomaly detection problem.

\textbf{Anomaly detection (prediction)}. The intuition behind the use of an autoencoder is that the loss of an anomalous instance would likely be much higher than that of a normal instance. The anomaly threshold is re-calculated every time the classifier is updated. At each training time $t$, we calculate the loss of all the elements in the queue $mov_{train}$: $L^t=\{l(x^i,\hat{x}^i)\}^t_{i=t-W_{train}+1}$. As shown in Eq.~(\ref{eq:threshold}), given a pre-specified value $b \in \{0, 1, .., 100\}$, the anomaly threshold at training time $t$ is calculated by taking the $b$, the percentile of the losses. For instance, if $b=90$, the threshold is set to the lowest value that is greater than 90\% of the losses. One advantage of the proposed approach is that the threshold is dynamic, and it can change at every training time.
\begin{equation}\label{eq:threshold}
\theta^t = percentile(b, L^t)
\end{equation}

Once having set a threshold, we can proceed with anomaly prediction. Suppose a new instance $x^{t+\Delta}$, $\Delta > 0$ is now observed. If its total loss is larger than the current threshold $\theta^t$, then it is classified as anomalous:
\begin{equation}\label{eq:predict}
\hat{y}^{t+\Delta} =
\begin{cases}
1 \ \text{(anomaly)} & if \ l(x^{t+\Delta}, \hat{x}^{t+\Delta}) > \theta^t\\
0 \ \text{(normal)} & \text{otherwise}
\end{cases}
\end{equation}
\noindent where $t$ is the time of the most recent training, and $t+\Delta$ is the current time.

\textbf{Incremental learning (training)}. The cost function $J^t$ at training time $t$ is defined as the average loss over all instances in the queue, shown in Eq.~(\ref{eq:cost}):
\begin{equation}\label{eq:cost}
J^t = \frac{1}{W_{train}} \sum_{i=t-W_{train}+1}^t l(x^i,\hat{x}^i)
\end{equation}
\noindent The autoencoder will be updated incrementally based on the cost incurred, that is, $h^t = h^{t-1}.train(J^t)$.

The frequency of updating the AE is critical. First, it would affect the computational cost, and secondly, it could determine the degree of overfitting. Therefore, training is performed when $p\%$ of the sliding window is replaced, as shown in Lines 14-15 in Algorithm 1. For instance, an expensive option would be to train the AE at each step upon observing the most recent instance, while if $p=50\%$, training will be performed after half of the window is replaced. After each incremental training, as shown in Line 16 in Algorithm 1, the losses and anomaly threshold will be re-calculated. The strAEm++'s part is shown in Algorithm 1 / Lines 10-16.

\subsection{StrAEm++DD}
\textbf{Memory.} This method uses a drift reference window $ref_{driftx}$ and a drift moving window $mov_{driftx}$ of size $W_{drift}$. The instances of drift reference window are obtained from the first window of arriving data and the drift moving window is used to store the most recent instances. These two windows are used for statistical comparison to detect concept drift. Another window $mov_{warn}$ of size $W_{drift}$ stores instances with warning flag, which are used to train the new AE and get the new threshold once the flag alarm is triggered.

The detection mechanism is realised by comparing the reconstruction loss of the reference window $ref_{driftl}$ and moving window $mov_{driftl}$ using the Mann–Whitney U Test. $ref_{driftl}$ and $mov_{driftl}$ are calculated by using $L^t=\{l(x^i,\hat{x}^i)\}^t_{i=t-W_{drift}+1}$. The corresponding part is from Line 17 to the end in Algorithm 1.

\textbf{Statistical test.} As a non‐parametric test, the Mann‐Whitney U Test does not depend on assumptions on the distribution\cite{nachar2008mann}. Benefiting from this property, researchers have applied this method to fault detection\cite{yang2019new},\cite{swain2019complete}. The test is performed by two hypotheses (H0 and H1).

H0: There is no difference between $ref_{driftl}$ and $mov_{driftl}$ (the two samples come from the same distribution).

H1: There is a difference between $ref_{driftl}$ and $mov_{driftl}$ (the two samples do not come from the same distribution).

The Mann-Whitney U Test's steps are described as follows.

The two group samples are put into one set first, and then sort the set in ascending order and assigned a numeric rank for each value of the set. Whenever there are ties between observations for the rank, we take the rank to be equal to the midpoint of the ranks. Then, we calculate the sum of the rank for the individual sample and get $R_{ref}$ and $R_{mov}$ for reference window and moving window.
Then we calculate the U value of reference window and moving window respectively as shown in Eq.~(\ref{eq:u})

\begin{equation}\label{eq:u}
\begin{aligned}
 U_{ref} &= n_{ref}\cdot n_{mov}+\frac{n_{ref}\left(n_{ref}+1\right)}{2}-R_{ref} \\
U_{mov} &= n_{ref}\cdot n_{mov}+\frac{n_{mov}\left(n_{mov}+1\right)}{2}-R_{ref}
\end{aligned}
\end{equation}

In Eq.~(\ref{eq:u}), $n_{ref}$ and $n_{mov}$ are sample sizes for reference window and moving windows, which is $W_{drift}$ in our case.

\begin{algorithm}[H]
	\caption{strAEm++DD}
	\label{alg:method}
	\begin{algorithmic}[2]
		
		\Statex \textbf{Input:} 
            \State $b$: percentile; $p$: window percentage replaced; $D$: unlabelled data for pre-training; $W_{drift}$: window size for drift detection; $W_{train}$: window size for re-training; $expiry\_time$: of the warning flag

		\Statex \textbf{Init:}\Comment time $t=0$
		\State $h = pretrain\_model(D)$
		\State {$mov_{train} (capacity=W_{train}, init=\{\})$}
		
		\State {$ref_{driftx} (capacity=W_{drift}, init=\{\})$} \Comment Ref. window (input)

		\State $mov_{driftx} (capacity=W_{drift}, init=\{\})$ \Comment Mov. window (input)
		\State $mov_{warn} (capacity=W_{drift}, init=\{\})$ \Comment window to store data with warning flag

        \State $flag_{warn} = False$ \Comment Drift warning flag
        \State $flag_{alarm} = False$ \Comment Drift alarm flag
        
		\State{$\theta = calc\_anomaly\_threshold(capacity=2000, init={D})$}

		\Statex \textbf{Main:}
		\For{each time step $t \in [1, \infty)$}
		\State receive instance $x^t \in \mathbb{R}^d$
		\State predict $\hat{y}^t = h.predict(x^t) \in \{0,1\}$
		\State append instance $mov_{train}.append(x^t)$

        	\If{ ($mov_{train}.isFull()$ or $p\%$ replaced) and $flag_{warn} == False$}\Comment \textcolor{blue}{Incremental learning}
        		    \State $h.train(mov_{train})$
        	        \State $\theta = calc\_anomaly\_threshold(mov_{train})$
        		\EndIf
            
            \If{$ref_{driftx}.isFull()$}\Comment \textcolor{blue}{DD mechanism}
                \State $ref_{driftl} = h.predict(ref_{driftx})$
  		    \State append instance $mov_{driftx}.append(x^t)$ 
            \Else
                \State append instance $ref_{driftx}.append(x^t)$
            \EndIf

    	\If {$mov_{driftx}.isFull()$}
    	        \State $mov_{driftl} = h.predict(mov_{driftx})$

                \If{$flag_{warn} == False$}
        		\If{$MWU(ref_{driftl}, mov_{driftl}) \leq P_{warn}$}
        		    \State $flag_{warn} = True$ \Comment Raise warning flag
        		\EndIf
                \EndIf
		
    		\If{$MWU(ref_{driftl}, mov_{driftl}) \leq P_{alarm}$}
    		    \State $flag_{alarm} = True$ \Comment Raise alarm flag
    		\EndIf
    	\EndIf

        \If {$flag_{warn}$ and $\neg flag_{alarm}$} \Comment \textcolor{blue}{Warning flag}
            \State append instance $mov_{warn}.append(x^t)$
            \If{flag raised for more than $expiry\_time$}
                \State $flag\_{warn} = False$
                \State $mov_{warn}$ empty
            \EndIf
        \EndIf

        \If {$flag_{alarm}$} \Comment \textcolor{blue}{Alarm flag}

    		\State $h = train(mov_{warn})$ \Comment new model with new AE
                \State $\theta = calc\_anomaly\_threshold(mov_{warn})$ 
                \State $ref_{driftx} (capacity=W_{drift}, init=\{\})$
    		\State $mov_{train}(capacity=W_{train},{init=\{\}}$ 
    		\State $mov_{driftx} (capacity=W_{drift}, init=\{\})$ 
    		\State $mov_{warn} (capacity=W_{drift}, init=\{\})$ 
                \State $flag_{warn} = False$
                \State $flag_{alarm} = False$
        \EndIf
        
	    \EndFor
    \end{algorithmic}
\end{algorithm}

\begin{equation}\label{eq:z}
Z=\frac{U-\mu_U}{\sigma_U}
\end{equation}
\noindent In Eq.~(\ref{eq:z}), $ \mu_U $ is the mean of U and $ \sigma_U $ is the standard deviation of U. The $ \mu_U $ and $ \sigma_U $ are calculated using Eq.~(\ref{eq:musigma})

The smallest value between $U_{ref}$ and $U_{mov}$ is used for calculation, i.e. $U=minimum(U_{ref} , U_{mov})$. Then we calculate Z-value using Eq.~(\ref{eq:z}).

\begin{equation}\label{eq:musigma}
\begin{aligned}
&\mu_U=\frac{n_{ref} \cdot n_{mov}}{2} \\
&\sigma_U=\sqrt{\frac{n_{ref} \cdot n_{mov}\left(n_{ref}+n_{mov}+1\right)}{12}}
\end{aligned}
\end{equation}

We will use this test to raise two flags $flag_{warn}$ and $flag_{alarm}$ as shown below. The warning flag indicates a warning for a potential concept drift, while the other flag triggers an alarm for an actual concept drift.
\begin{equation}\label{eq:pvalue}
{flag} =
\begin{cases}
warn \  & if \ P_{value} < P_{warn}\\
alarm \ &  if \ P_{value} < P_{alarm}\\
\end{cases}
\end{equation}

The P-value is calculated by using the Z-value for the standard normal distribution table. The P-value is the two-tailed standard normal probability with $\alpha=0.05$ significance. The P-value is defined in Eq.~(\ref{eq:pvalue_def}), where $\mu$ is the mean and $\sigma^2$ is the variance of the observation $Z(t)$. $P_{warn}$ and $P_{alarm}$ are threshold values set for warning and alarm. The P-value for $flag_{warn}$ should be larger than the p-value for $flag_{alarm}$, i.e., $P_{warn} > P_{alarm}$. Once there is a flag alarm, then H0 (null hypothesis) is rejected and H1 (alternative hypothesis) is satisfied. The above description is shown in Lines 24 - 28 in Algorithm 1.

\begin{equation}\label{eq:pvalue_def}
P_t=2 \int_{-\infty}^t \frac{1}{\sqrt{2 \pi \sigma^2}} \times e^{\left(-\frac{Z(t)-\mu}{2 \sigma^2}\right) d t}, \sigma>0
\end{equation}

\textbf{Warning flag raised}. As shown in Lines 29-33 in Algorithm 1, once a $flag_{warn}$ is raised, we start to store examples into $mov_{warn}$ and pause incremental learning. In order to avoid false alarm, we set a parameter $expiry\_time$, if $flag_{warn}$ is raised for more than $expiry\_time$ and there is still no $flag_{alarm}$, we regard this as false warning and then reset the status of $flag_{warn}$ and empty $mov_{warn}$.

\textbf{Alarm flag raised}. When a $flag_{alarm}$ is raised, we create a new autoencoder to replace the old one and train this new model with $mov_{warn}$. And also, the threshold is updated with $mov_{warn}$. And all the windows, $ref_{driftx}$, $mov_{train}$, $mov_{driftx}$ and $mov_{warn}$ are emptied and two flags are reset. The above steps can be found in Lines 34-42 in Algorithm 1. The new reference window after drift will be refilled by arriving instances of size $W_{drift}$ after drift as demonstrated back to Line 21 in Algorithm 1.

\section{Experimental Setup}\label{sec:exp_setup}

\subsection{Datasets}

Our experimental study considers: (i) synthetic (Sea, Circle) and real-world (MNIST-01, MNIST-23) datasets, (ii) different types of concept drift (change in the posterior or class-conditional probability), and (iii) imbalance cases with imbalance rates of 1\% (severe) and 0.1\% (extreme).

\textbf{Sea}: It has two features $x_1, x_2 \in [0,10]$. Before drift, instances that satisfy $x_1 + x_2 \leq 7$ are classified as positive, otherwise as negative. After the drift at $t=5000$, the posterior probability $p(y|x)$ of the data distribution alters abruptly, i.e, the normal class swaps with the anomalous class. Re-scaling has been performed so that $x_1, x_2 \in [0,1]$.

\textbf{Circle}: It has two features $x_1, x_2 \in [0,1]$. The decision
boundary is a circle with center $(0.4, 0.5)$ and radius $0.2$. Examples that fall inside the circle are classified as positive $(y = 1)$ and outside as negative $(y = 0)$. The posterior probability $p(y|x)$ of the data distribution alters abruptly after the drift at $t=5000$. The normal class swaps with the anomalous class.

\textbf{MNIST} \cite{lecun1998gradient}: It consists of handwritten images of digits, each of size 28x28. While it, typically, serves as a benchmark dataset in image classification systems, it can challenge streaming methods with its high dimensionality of 784 features. We consider two variations as follows:

\begin{itemize}
    \item \textbf{MNIST-01}: It consists of 5000 images of digits ``0'' and ``1'', Digit ``1'' is considered to be the minority (anomalous) class. The concept drift in this dataset occurs at $t=2500$, where the width and height of images are shifted by +/-10\%. This drift causes a change in the class-conditional probability or likelihood $p(x|y)$.

    \item \textbf{MNIST-23} : It consists of 5000 images of digits ``0'' - ``3''. Before the drift at $t=2500$, the majority (normal) class is the digit ``0'' and the minority (anomalous) class is the digit ``1''. After the drift, the majority class is the digit ``2'' and the minority class is the digit ``3''. The drift causes a change in the posterior probability $p(y|x)$. 
\end{itemize}

\subsection{Methods}

\textbf{Baseline}: In this method we simply perform pre-training. The training of this method is offline with 2000 unlabelled examples of normal class. The baseline method is also applied in the other methods below as the pre-trained model.

\textbf{strAEm++}: It corresponds to incremental learning with Autoencoder without drift detection mechanism. This is the proposed method as described in Section IV-A.

\textbf{iForest++}: A state-of-the-art tree-based method \cite{liu2008isolation} described in Section III-B. For fairness, the same framework is used as shown in Fig. 1 and described in section IV-A but with an Isolation Forest (iForest) instead of an AE.

\textbf{strAEm++DD}: The proposed method as described in Section IV-B and its pseudocode is shown in Algorithm~\ref{alg:method}.

To facilitate the reproducibility of our results, Table~\ref{tab:params_nn} provides the hyper-parameters for strAEm++(DD) for all datasets where fully connected neural networks are used. Table~\ref{tab:params_iforest} provides the relevant information for iForest++.

\begin{table}
\caption{Hyper-parameter values for strAEm++(DD)}\label{tab:params_nn}
\centering
\begin{tabular}{|c|c|c|c|} 
\hline
                   & Sea      & Circle          & MNIST         \\ 
\hline
Learning rate      & \multicolumn{2}{c|}{0.001} & 0.0001        \\ 
\hline
Hidden layers      & {[}64,8] & 8               & {[}512,256]   \\ 
\hline
Mini-batch size    & \multicolumn{3}{c|}{128}                   \\ 
\hline
Weight initializer & \multicolumn{3}{c|}{He Normal}             \\ 
\hline
Optimizer          & \multicolumn{3}{c|}{Adam}                  \\ 
\hline
Hidden activation  & \multicolumn{3}{c|}{Leaky ReLU}            \\ 
\hline
Num. of epochs     & 10       & 5               & 10            \\ 
\hline
Output activation  & \multicolumn{3}{c|}{Sigmoid}               \\ 
\hline
Loss function      & \multicolumn{3}{c|}{Binary cross-entropy}  \\
\hline
\end{tabular}

\end{table}

\begin{table}[]
\caption{Hyper-parameter values for iForest++}\label{tab:params_iforest}
\centering
\begin{tabular}{|c|ccc|}
\hline
                         & \multicolumn{1}{c|}{Sea} & \multicolumn{1}{c|}{Circle} & MNIST \\ \hline
Num. of estimators       & \multicolumn{3}{c|}{100}                                       \\ \hline
Maximum num. of features & \multicolumn{2}{c|}{2}                                 & 784   \\ \hline
Maximum num. of samples  & \multicolumn{3}{c|}{256}                                       \\ \hline
Proportion of outliers   & \multicolumn{3}{c|}{0.1}                                       \\ \hline
\end{tabular}

\end{table}

\subsection{Performance metrics}
One suitable and widely accepted metric that is insensitive to class imbalance is the geometric mean \cite{sun2006boosting}, defined as:
\begin{equation}\label{eq:gmean}
G\text{-}mean = \displaystyle \sqrt{R^+ \times R^-},
\end{equation}

\noindent where $R^+ = TP / P$ is the recall of the positive class, $R^-=TN / N$ is the recall (or specificity) of the negative class, and TP, P, TN, and N are the number of true positives, total positives, true negatives, and total negatives, respectively.

\noindent Not only G-mean is insensitive to class imbalance, it has some important properties as it is high when all recalls are high and when their difference is small \cite{he2008learning}.

A popular method to evaluate sequential learning algorithms is the \textit{prequential evaluation with fading factors} method. It has been proven to converge to the Bayes error when learning in stationary data \cite{gama2013evaluating}. Its major advantage is that it does not require a holdout set and the classifier is always tested on unseen data. We set the fading factor to $\xi = 0.99$.

For statistical significance, in all simulation experiments we plot the prequential metric (i.e., G-mean) in every time step averaged over 20 repetitions, including the error bars displaying the standard error around the mean.

\section{Experimental Results}\label{sec:exp_results}

\subsection{Empirical analysis of strAEm++}

 We conduct experiments with different window sizes and different number of epochs for training to investigate the influence of these parameters on the model performance.

\subsubsection{Role of the window size}
Fig. 2 shows the model’s performance for different window sizes. At the beginning, we set the percentile of the losses as $b=95$ and the number of epochs as 10. Then we conduct experiments with different window size $W_{train}$ (100, 500, 800, 1000, 2000).
As shown in Fig. 2, for datasets with both severe and extreme imbalance, the variation of window size doesn't make much difference in performance before the drift because the model has been pre-trained. After the drift, we can observe that with the largest window $W_{train}=2000$, it takes much longer to adapt to the concept drift compared to the smallest window $W_{train}=100$, but its performance is eventually much better.

\begin{figure}[t!]
\begin{subfigure}{.5\columnwidth}
  \centering
  \includegraphics[width=.9\columnwidth]{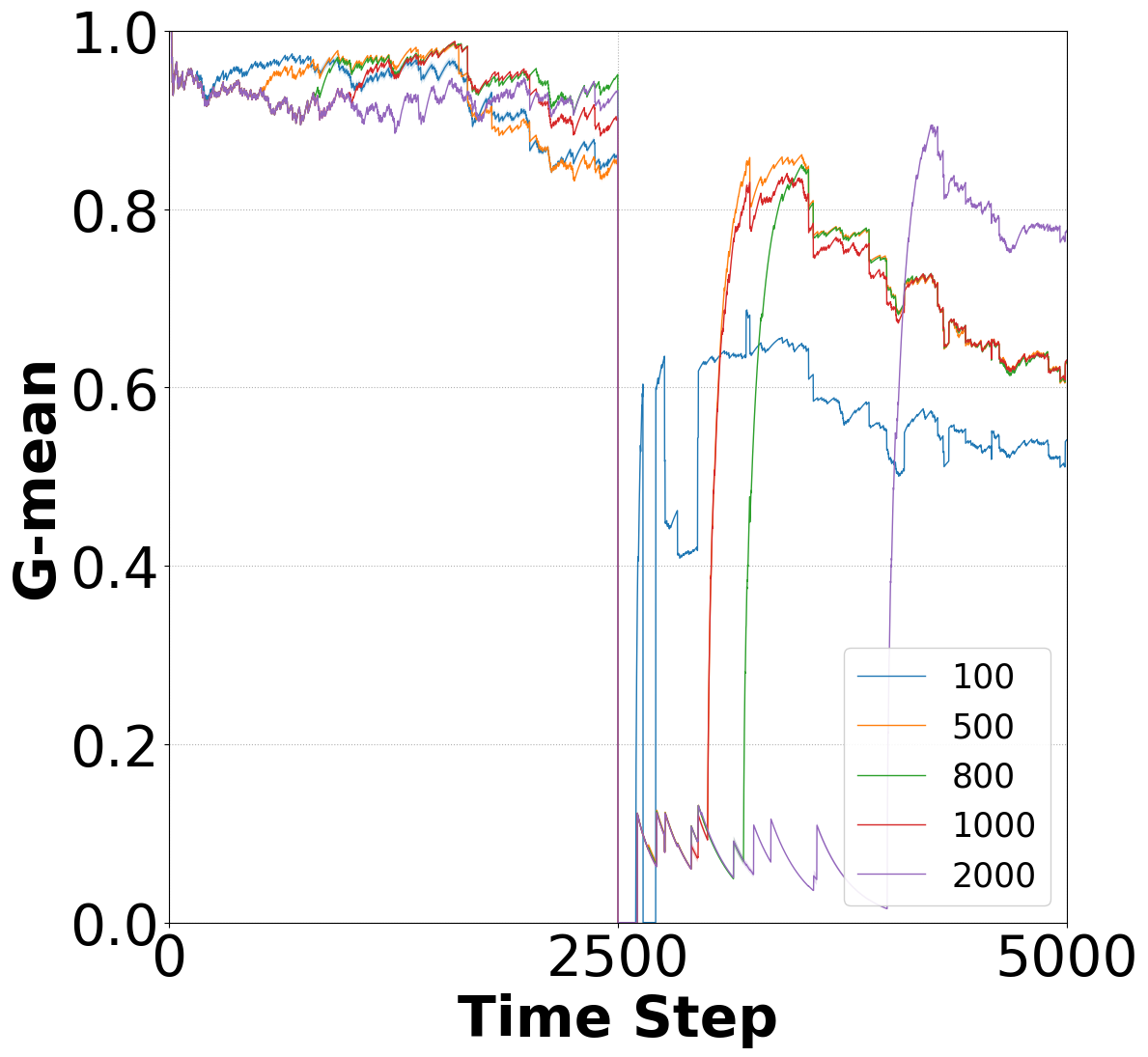}
  \caption{MNIST-23 severe}
  \label{fig:sfig_mnist23_severe_win}
\end{subfigure}%
\begin{subfigure}{.5\columnwidth}
  \centering
  \includegraphics[width=.85\columnwidth]{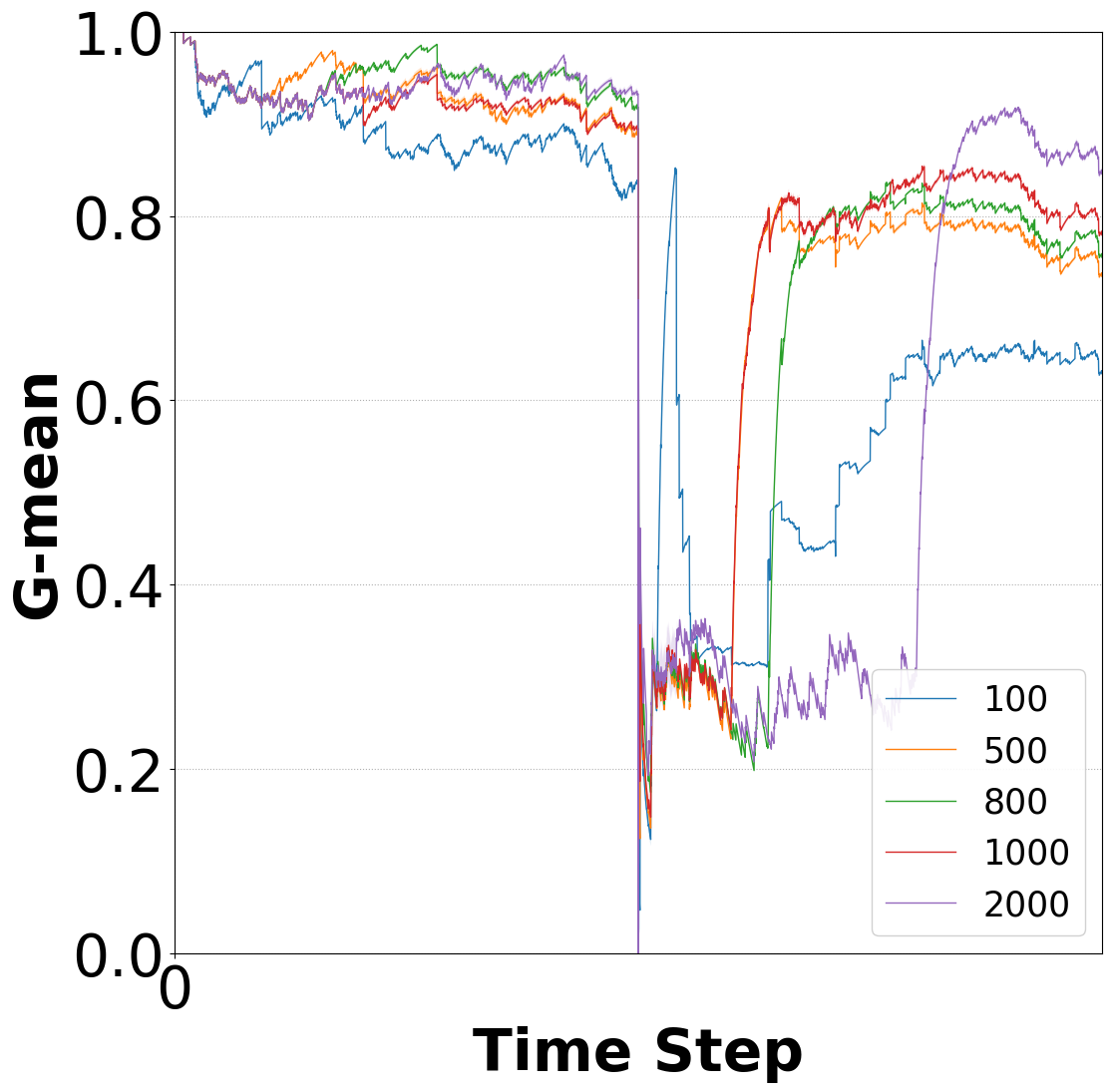}
  \caption{MNIST-01 severe}
  \label{fig:sfig_mnist_m01_severe_win}
\end{subfigure}
\caption{ Performance of strAEm++ in non-stationary environments with different window sizes.}
\end{figure}
Overall, the results indicate the tendency that bigger window size leads to slower drift adaptation but with better performance. However, it can be also seen that the performance after drift is not totally recovered, for which we need to tune more parameters to see if it can be improved.

\subsubsection{Role of the number of epochs}
From the previous results, we select window size $W_{train}=1000$ and $W_{train}=100$ for MNIST-23 and MNIST-01 respectively, with which the model performs the best in Fig. 2. Then we conduct experiments with the selected window size and different number of epochs (1, 10, 50, 100, 300, 500). The percentile of the losses is still fixed to $b=95$. 

\begin{figure}[t!]
\begin{subfigure}{.5\columnwidth}
  \centering
  \includegraphics[width=0.9\columnwidth]{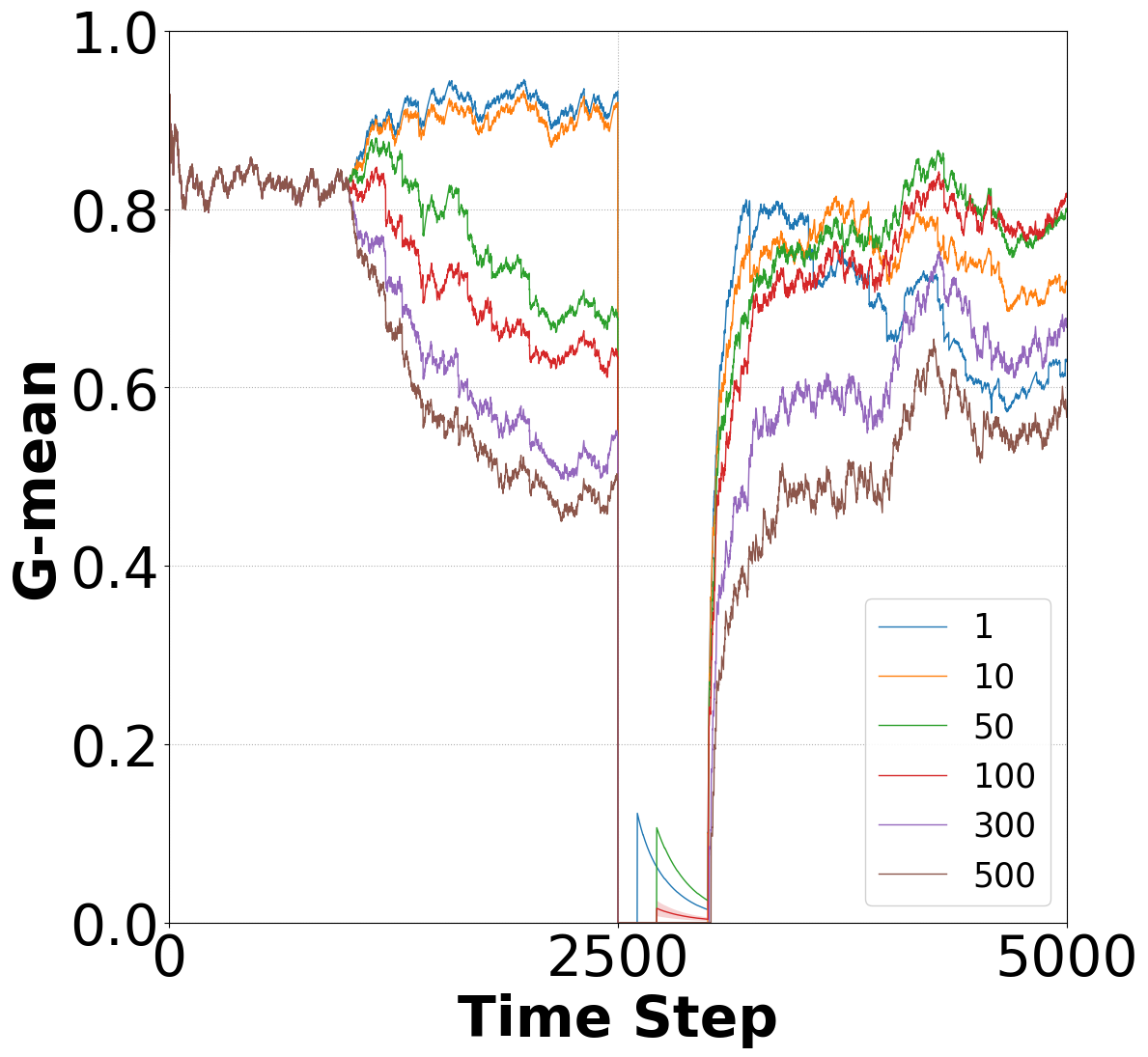}
  \caption{MNIST-23 severe}
  \label{fig:sfig_mnist23_severe_epoch}
\end{subfigure}%
\begin{subfigure}{.5\columnwidth}
  \centering
  \includegraphics[width=.9\columnwidth]{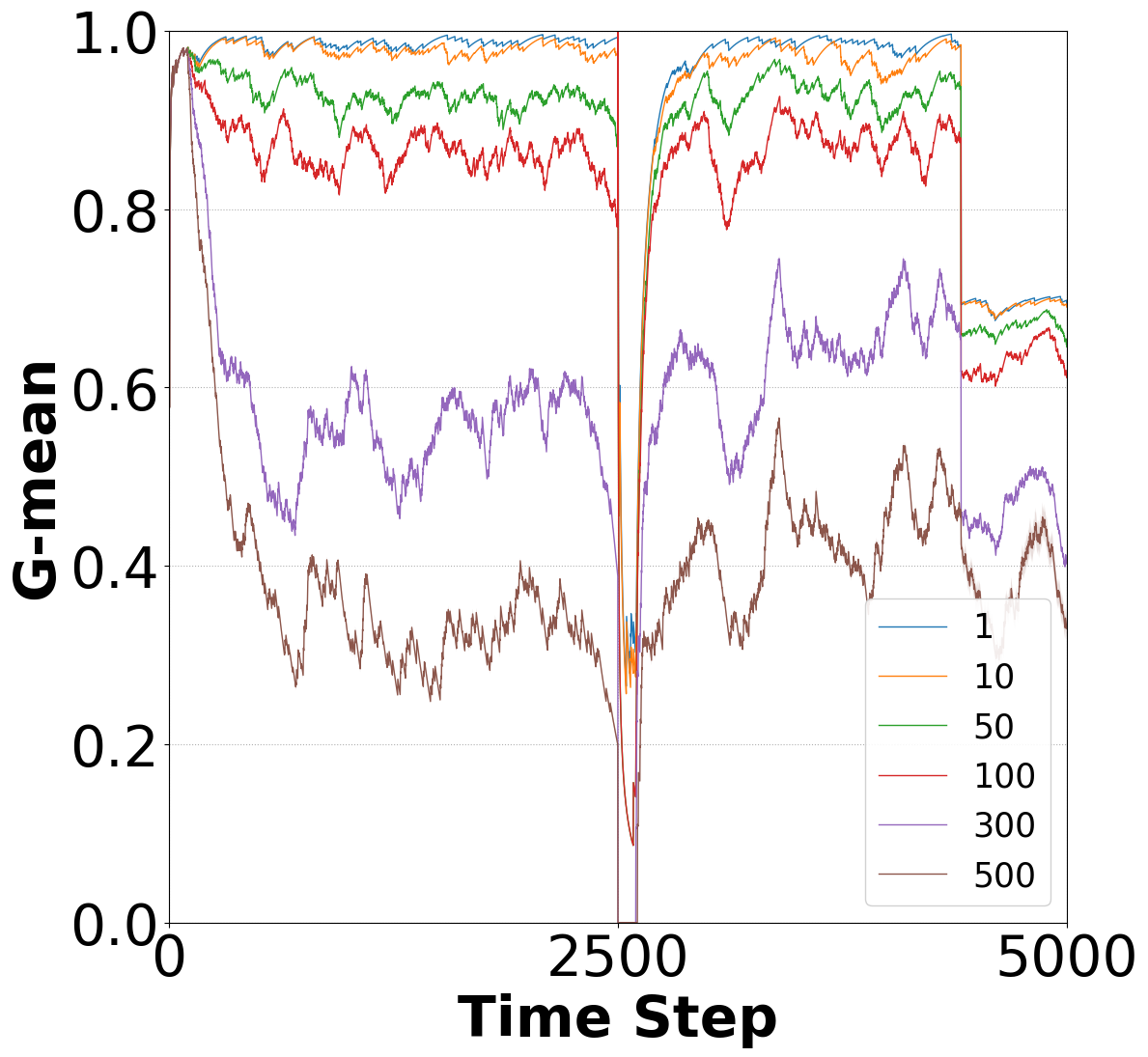}
  \caption{MNIST-01 extreme}
  \label{fig:sfig_mnist_01_extreme_epoch}
\end{subfigure}
 \caption{ Performance of strAEm++ in non-stationary environments with different number of epochs.}
\end{figure}
 
As shown in Fig. 3, before the drift, the performance is inversely proportional to the number of epochs for both imbalanced cases. After the drift, a smaller number of epochs such as 1 or 10 still outperform the rest, which indicates that a small number of epochs suits more our model. This is attributed to the fact that a smaller number of epochs (e.g., up to 10) does not allow the model to overfit the most recently observed instances, thus achieving better generalisation.

\subsection{Empirical analysis of strAEm++DD}

We conduct experiments with different sizes of the reference and moving windows, and different values of $P_{warn}$ to investigate the influence of these
parameters on the model performance and the number of drift warnings respectively.

\subsubsection{Role of the size for reference / moving windows}
The dataset MNIST-23 under severe imbalance is employed in this experiment. We set $expiry\_time=100$, $W_{train}=1000$, $P_{warn}=0.07$ and $P_{alarm}=0.001$ and we compare the model performance and the number of warnings with different reference / moving window size $W_{drift}$. Fig.4 (a) shows that $W_{drift}=200$ and $W_{drift}=300$ have the best performance and, as shown in Fig.4 (b), and they both have three error warnings, less than the other two windows. Therefore, we will take $W_{drift}=200$ for the next experiment to investigate the role of $P_{warn}$.
\subsubsection{Role of the P-value}
In this experiment, we only investigate the role of $P_{warn}$. $P_{alarm}$ is fixed to the small value 0.001, to minimize the possibility of false alarms. As can be seen from Fig.5 (a), the variation of value $P_{warn}$ merely affects the performance of model. From Fig.5 (b), we can see that the number of false warnings increases with the increasing $P_{warn}$. It is reasonable according to the Eq.~(\ref{eq:pvalue}). When $P_{warn}$ is larger, more instances would be regarded as warnings. 
\begin{figure}[H]
\begin{subfigure}{.5\columnwidth}
  \centering
  \includegraphics[width=.9\columnwidth]{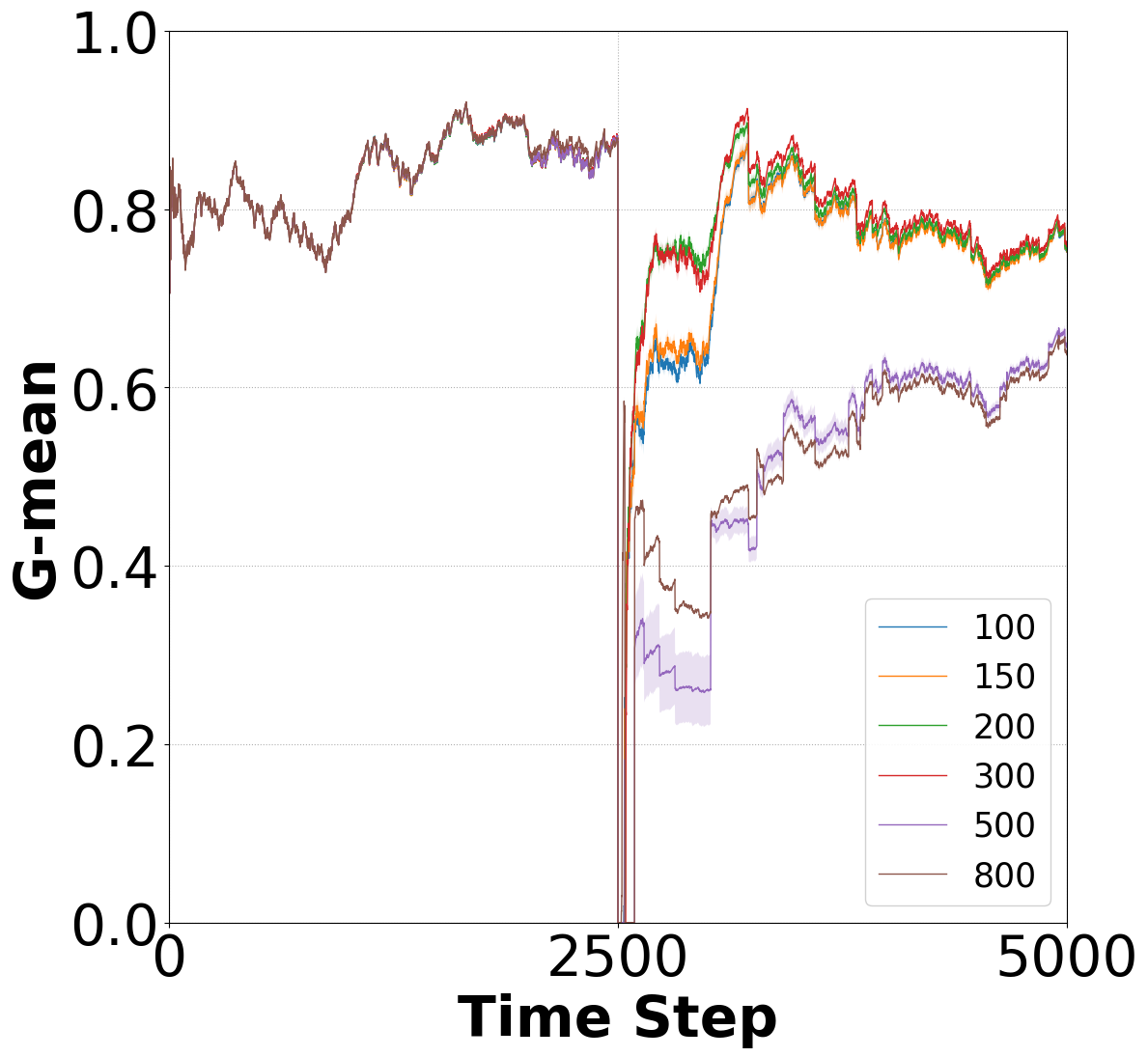}
  \caption{model performance}
  \label{fig:mnist_23_severe_performance_diffwin}
\end{subfigure}%
\begin{subfigure}{.46\columnwidth}
  \centering
  \includegraphics[width=.9\columnwidth]{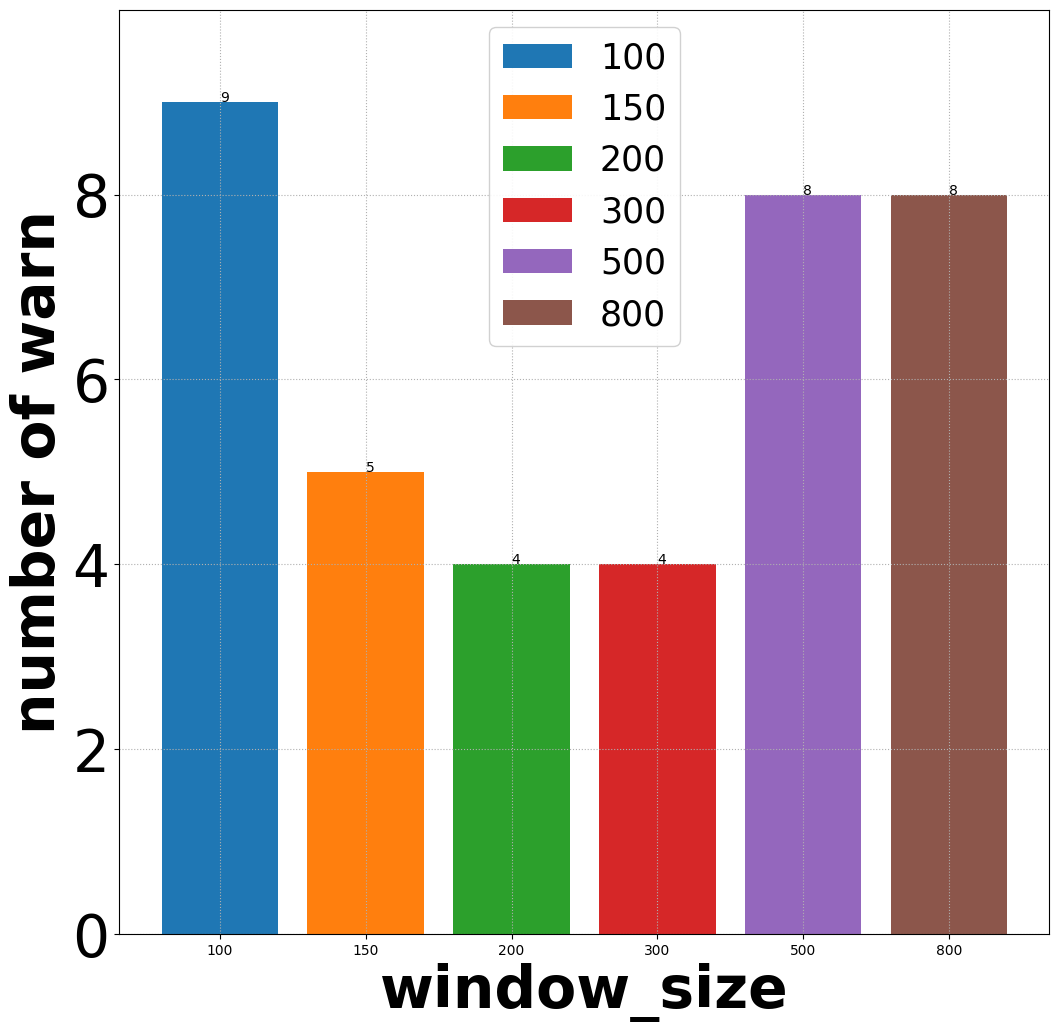}
  \caption{number of warnings}
  \label{fig:sfig_mnist_23_extreme}
\end{subfigure}
\caption{Performance and number of warnings of strAEm++DD in non-stationary environments with different window size and dataset MNIST-23 severe}
\label{fig:fig_mnist_23}
\end{figure}

\begin{figure}[H]
\begin{subfigure}{.5\columnwidth}
  \centering
  \includegraphics[width=.9\columnwidth]{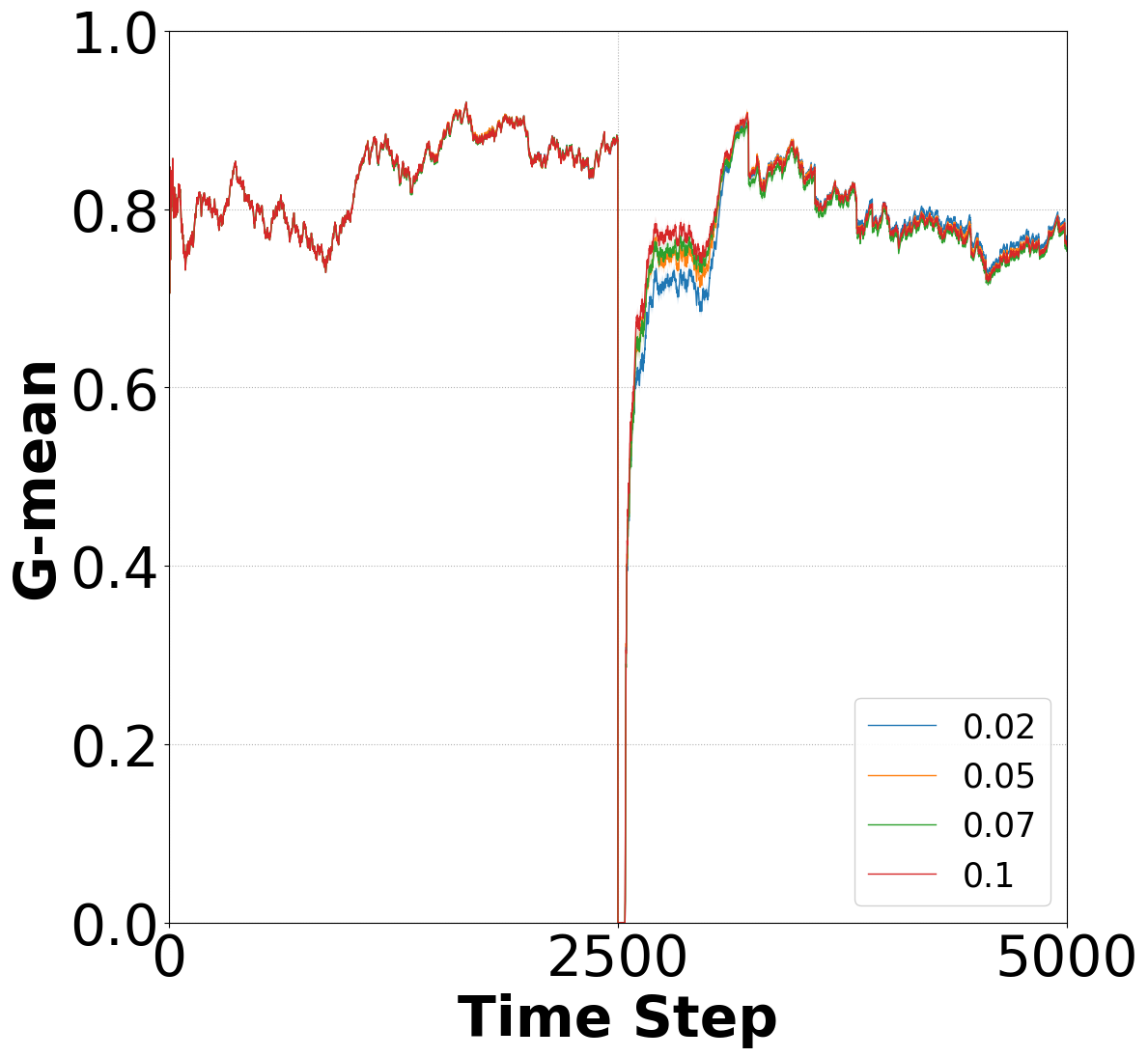}
  \caption{model performance}
  \label{fig:mnist_23_severe_performance_diff}
\end{subfigure}%
\begin{subfigure}{.46\columnwidth}
  \centering
  \includegraphics[width=.9\columnwidth]{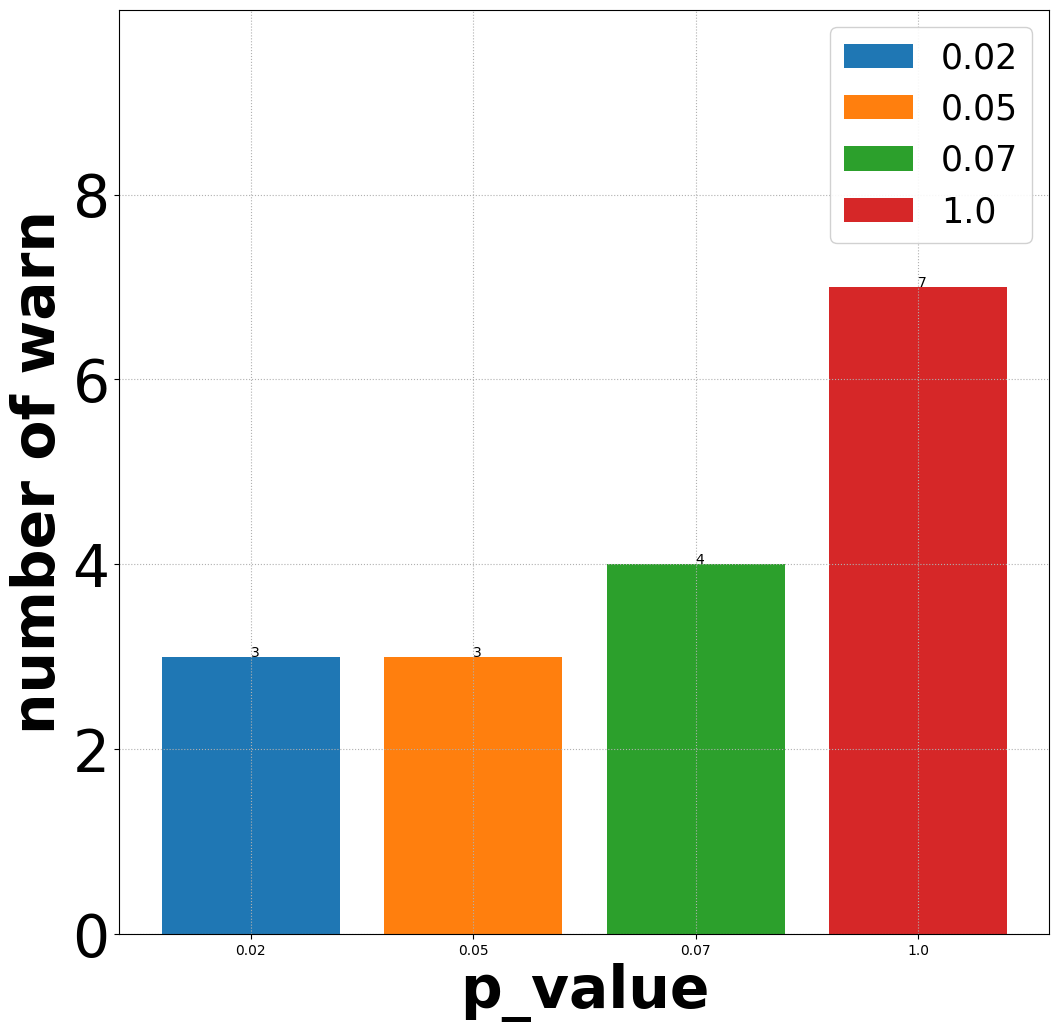}
  \caption{number of warnings}
  \label{fig:sfig_mnist_23_extreme}
\end{subfigure}
\caption{Performance and number of warnings of strAEm++DD in non-stationary environments with difference P-values and dataset MNIST-23 severe}
\label{fig:fig_mnist_23}
\end{figure}

\subsection{Comparative study}

In this section we compare the performance of four different methods: Baseline, strAEm++, iForest++ and strAEm++DD. The detailed descriptions of these methods are shown in Section IV. For datasets MNIST-01, MNIST-23 and Sea, the parameters are set as $W_{train}=1000$, $W_{drift}=200$, $b=80$, $P_{warn}=0.01$, $P_{alarm}=0.001$, and the number of epochs to 10. For dataset Circle, all parameters are the same except the number of epoch is set to 5.

The results are shown in Fig. 6. A reasonable number of false drift warnings, and only one drift alarm are shown in all plots, which proves the good detectability and robustness of our proposed method strAEm++DD. As it can be seen in Fig.6, before the drift, the G-mean value of strAEm++ and strAEm++DD is always above 0.8; while in Fig.6 (d), the performance of iForest++ is significantly lower. In Fig.6 (b), Fig.6 (c) and Fig.6 (d), after drift, the proposed method strAEm++DD outperforms strAEm++, which proves the necessity of drift detection mechanism in non-stationary dataset. Apart from that, in these three sub-figures, strAEm++DD clearly outperforms iForest++ for all time steps or 2000 time steps after drift as shown in Fig.6 (c), which shows the advantage of strAEm++DD over iForest++.

In Fig.6 (a), strAEm++ and iForest++  show better performance than strAEm++DD. We attribute the performance difference to the drift type of each dataset. The difference between the instances before and after drift in datasets MNIST-23, Sea and Circle are more significant than that in dataset MNIST-01. Therefore, when the drift happens, the training memory of the classifier for dataset MNIST-01 can help to obtain a good performance. However, for strAEm++DD, it needs more time and instances to reach such a performance.

\begin{figure}[t!]
 \begin{subfigure}{.5\columnwidth}
  \centering
  \includegraphics[width=0.9\columnwidth]{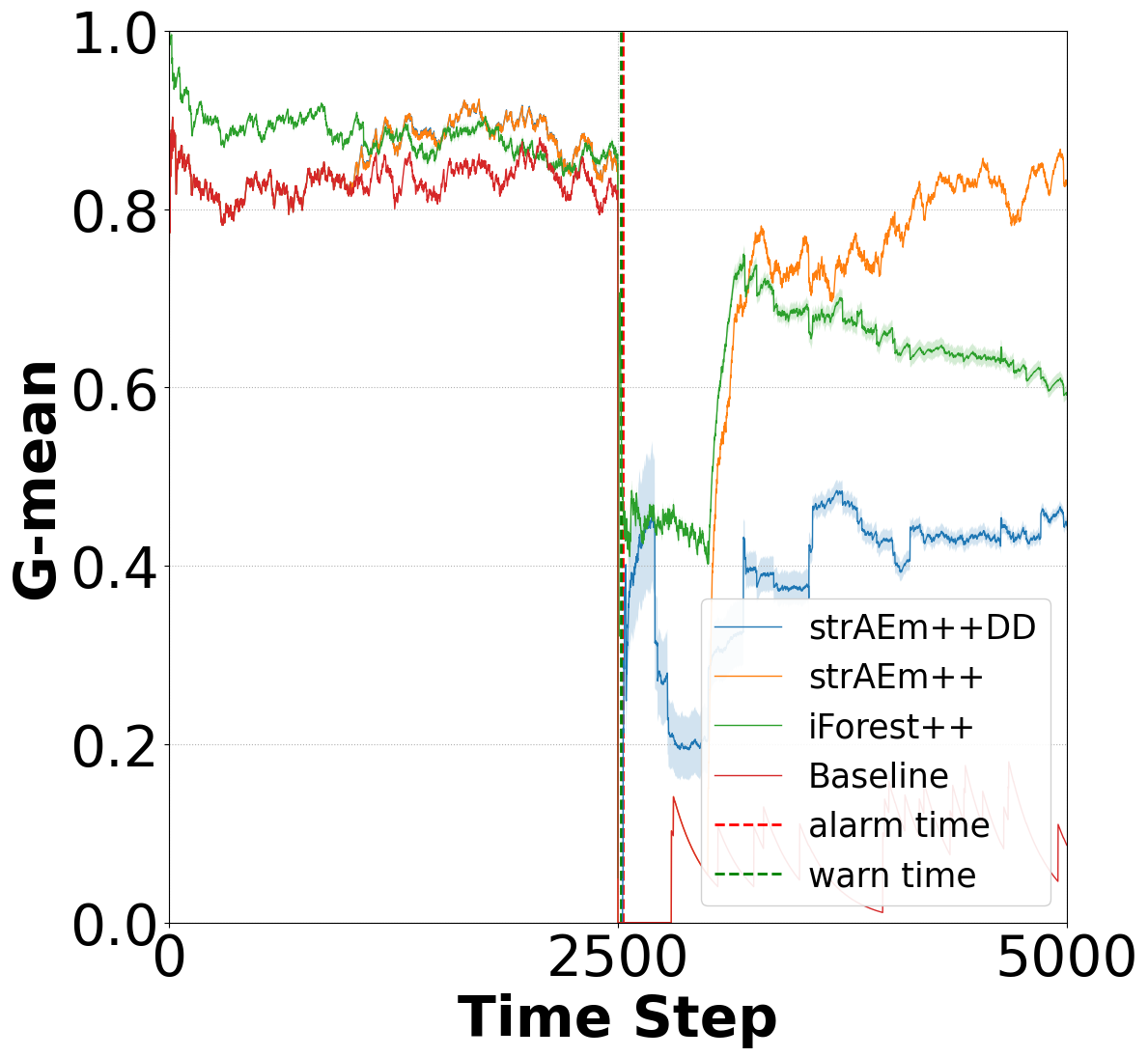} 
  \caption{MNIST-01 severe}
  \label{fig:mnist01_severe_compare}
 \end{subfigure}%
\begin{subfigure}{.5\columnwidth}
  \centering
  \includegraphics[width=0.9\columnwidth]{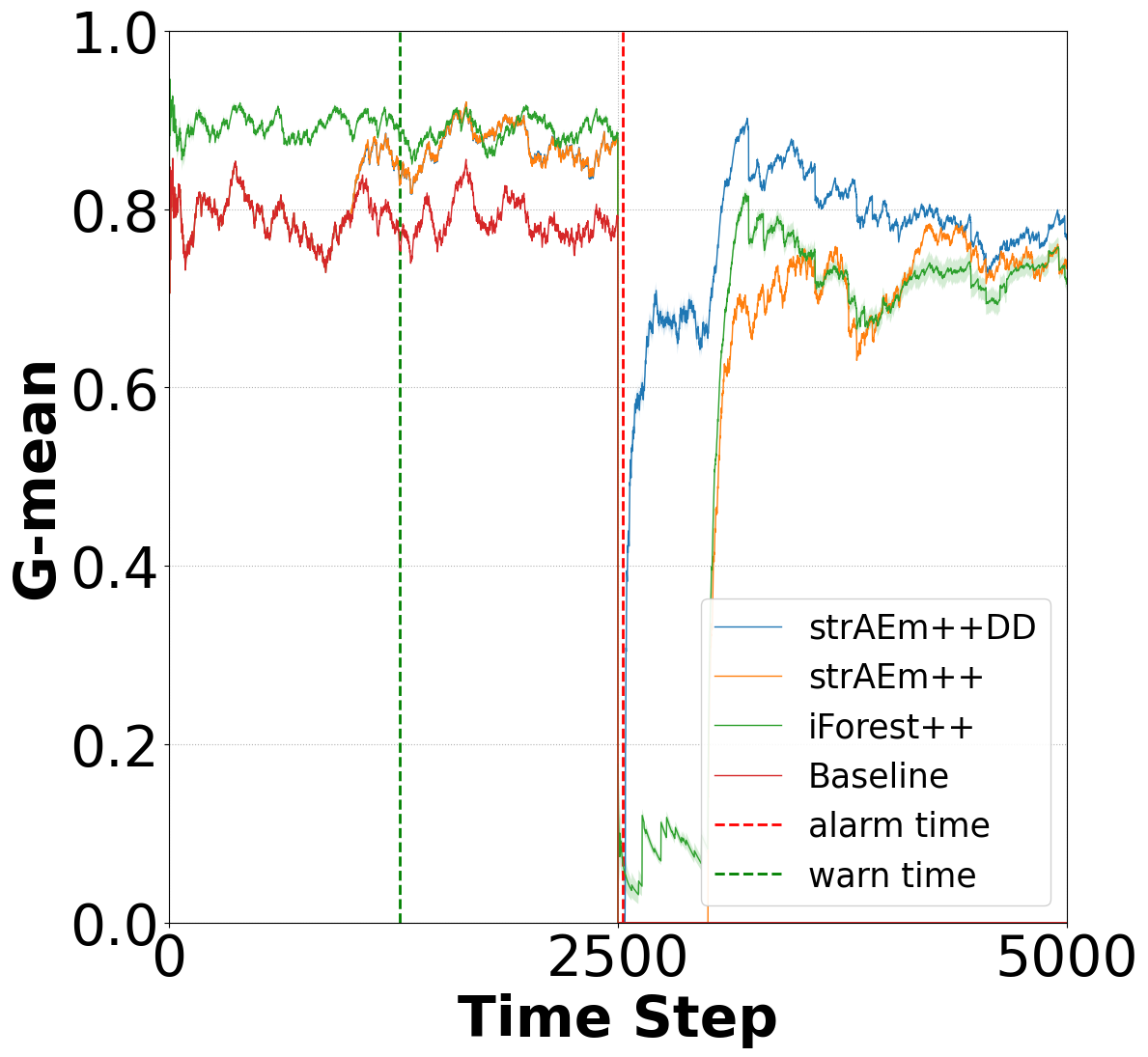}
  \caption{MNIST-23 severe}
  \label{fig:mnist23_severe_compare}
\end{subfigure}%

 \begin{subfigure}{.5\columnwidth}
  \centering
  \includegraphics[width=0.9\columnwidth]{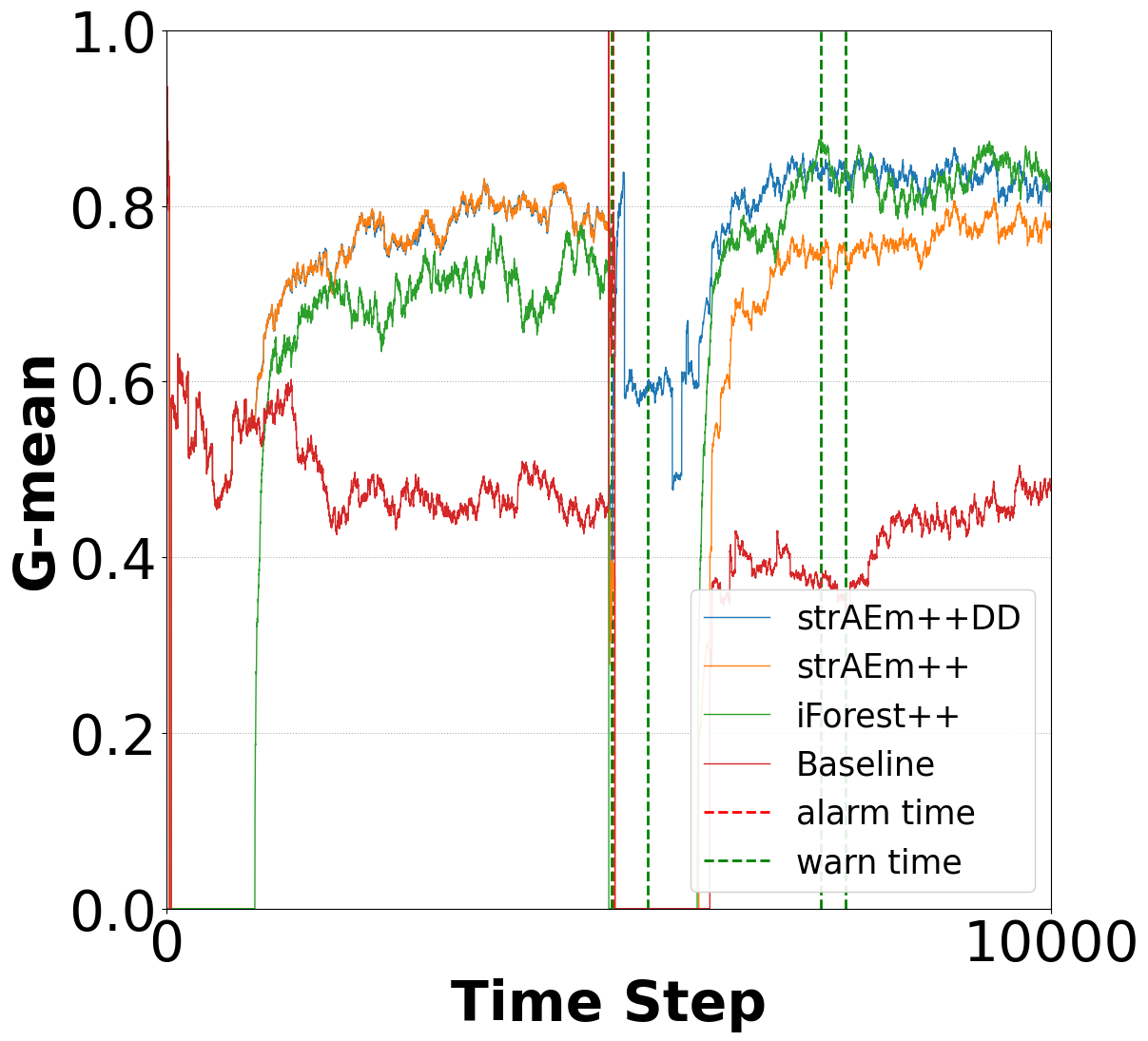} 
  \caption{Sea severe}
  \label{fig:Sea_severe_compare}
 \end{subfigure}%
 \begin{subfigure}{.5\columnwidth}
  \centering
  \includegraphics[width=0.83\columnwidth]{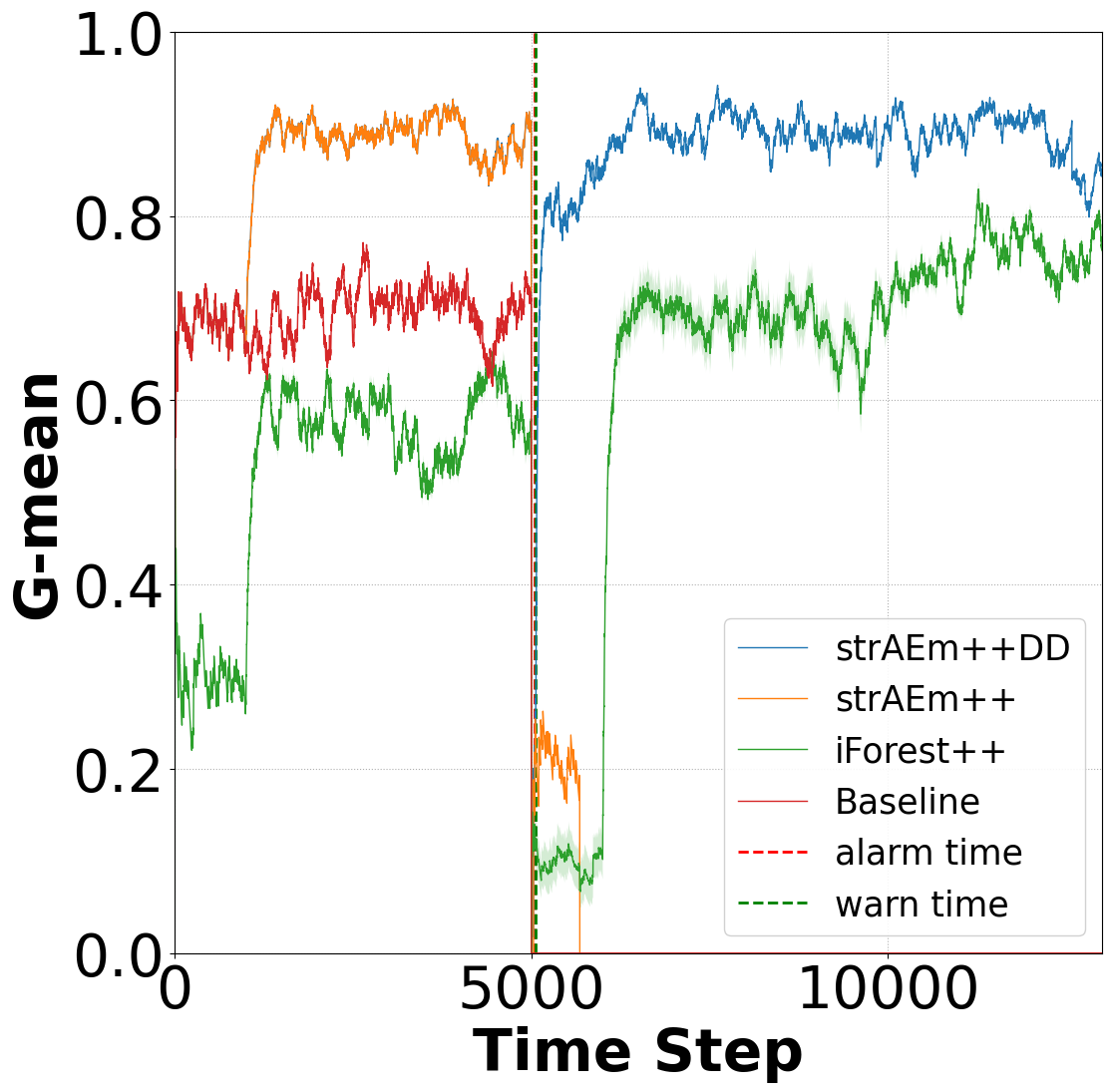} 
  \caption{Circle extreme}
  \label{fig:Circle_extreme_compare}
 \end{subfigure}%
\caption{Comparison between Baseline, strAEm++, iForest++ and strAEm++DD in nonstationary environments.}
\label{fig:fig_compare}
\end{figure}

Overall, important remarks are as follows:
\begin{itemize}
        \item The proposed method strAEm++DD has a certain robustness with imbalanced classes.
	\item Generally, strAEm++DD outperforms strAEm++, iForest++ and Baseline methods. However, this also depends on the drift type. For drifts causing a change in the posterior probability $p(x|y)$, strAEm++DD has a more significant advantage over incremental methods strAEm++ and iForest++.
\end{itemize}

\section{Conclusion}\label{sec:conclusion}
Extracting patterns from data streams poses significant challenges, which includes the identification of infrequent events, ground truth unavailability, as well as adapting to nonstationary environments. To address the above challenges, we have proposed a novel method called strAEm++DD, which is an autoencoder-based incremental learning method with drift detection. Not only the application of autoencoders is limited within the online learning framework, but this is one of the very few studies that considers a hybrid active - passive approach to address concept drift. We have conducted an extensive experimental study in which we demonstrate that the proposed method significantly outperforms existing baseline and advanced methods. For future work, we will investigate different types of AEs (e.g., variational), ensemble learning, as well as applying the proposed method to more complex datasets.

\bibliographystyle{IEEEtran}
\bibliography{paper}

\end{document}